# Comparison of Reinforcement Learning Algorithms applied to the Cart-Pole Problem

Savinay Nagendra
PES Center for Int. Sys.
Dept. of EEE, PESIT
Bangalore, India
nagsavi17@gmail.com

Nikhil Podila
PES Center for Int. Sys.;
Dept. of EEE, PESIT
Bangalore, India
nikhilpodila.94@gmail.com

Rashmi Ugarakhod
PES Center for Int. Sys.;
Dept of ECE, PES University
Bangalore, India.
rashmi.ugarakhod@gmail.com

Koshy George
PES Center for Int. Sys.;
Dept. of Tel. Engg.,
PESIT, Bangalore, India.
kgeorge@pes.edu

*Abstract*—Designing optimal controllers continues to be challenging as systems are becoming complex and are inherently nonlinear. The principal advantage of reinforcement learning (RL) is its ability to learn from the interaction with the environment and provide optimal control strategy. In this paper, RL is explored in the context of control of the benchmark cart-pole dynamical system with no prior knowledge of the dynamics. RL algorithms such as temporal-difference, policy gradient actor-critic, and value function approximation are compared in this context with the standard LQR solution. Further, we propose a novel approach to integrate RL and swing-up controllers.

## I. INTRODUCTION

Reinforcement learning (RL) is a branch of machine learning which is inspired by human and animal behaviorist psychology. When RL is applied to a system, the agents of the system learn to take actions in an environment so as to maximize some notion of cumulative reward. Learning can be based on several forms of evaluative feedback [1], [2]. In contrast to supervised learning methods, RL is used when the target outputs are not known. Here, the performance is evaluated indirectly by considering the effect of the output on the environment with which the system interacts; this effect is quantified in terms of an evaluation or reinforcement signal.
RL algorithms focus on online performance, which involves finding a balance between *exploration* (of uncharted territory) and *exploitation* (of current knowledge). To obtain maximum reward, the agent has to exploit what it already knows, but it also has to explore in order to make better action selections in the future. Neither exploration nor exploitation can be pursued exclusively without failing at the task.

The cart-pole problem is a classical benchmark problem for control purposes. It is an inherently unstable and under-actuated mechanical system. The dynamics of this system is used to understand tasks involving the maintenance of balance, such as walking, control of rocket thrusters and self-balancing mechanical systems. A number of control design techniques for swing-up and stabilization of an inverted pendulum have been investigated. Examples include energy-based controllers, PID controllers, Linear Quadratic Regulators (LQR), and Fuzzy logic controllers; e.g. [3], [4].

An increase in the complexity of systems requires the need for sophisticated controllers especially in the presence of nonlinearities, uncertainty and time-variations. By its inherent nature, RL has the capability to use knowledge from the environment to provide optimal controllers without the knowledge of the environment. Moreover, such controllers have the capability to adapt to a changing environment.

The goal of this paper is two-fold. First, we compare several RL algorithms in the context of the cart-pole problem. We further assess which of these algorithms provide a control that resembles closer to the LQR solution. Secondly, we propose a method to integrate RL algorithm with a swing-up controller.

The rest of the paper is organized as follows: The cart-pole problem is described in Section II. The different RL algorithms that are of interest in this paper are presented in Section III. The classical swing-up control together with LQR stabilization is dealt with in Section IV. The manner in which RL algorithm is integrated with a swing-up controller is given in Section V. The performances of these controllers are compared in Section VI.

## II. CART-POLE PROBLEM

The cart-pole balancing problem is a benchmark for RL algorithms; e.g., [5]-[8]. The fundamental problem statement has been derived from an adaptive control technique known as the BOXES [8]. The problem is challenging as the reinforcement learning agent has to select and take actions in a very limited and discrete action space.

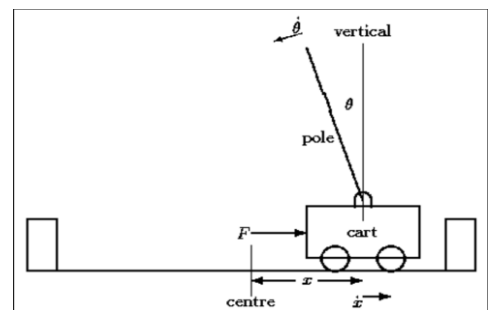

Fig 1. Cart-Pole system

### A. Cart-Pole Dynamics

A pendulum is pivoted to a cart, which has one degree of freedom along the horizontal axis. The goal of the problem is to balance the pendulum in the upright position by using bi-directional forces that is imparted on the cart by an electromechanical system. The state of the system at any time is defined by the angular position $\theta$ and velocity $\dot{\theta}$ of the pendulum, and the linear position $x$ and velocity $\dot{x}$ of the cart.

The nonlinear system is described by the following equations:

$$\ddot{\theta} = \frac{(M+m)g\sin\theta - \cos\theta[F + ml\dot{\theta}^2 \sin\theta]}{\left(\frac{4}{3}\right)(M+m)l - ml\cos^2\theta} \qquad (1)$$

$$\ddot{x} = \frac{\{F + ml[\dot{\theta}^2 \sin\theta - \ddot{\theta}\cos\theta]\}}{(M+m)} \qquad (2)$$

Here, M is the mass of the cart (0.711 kg), m is the mass of pendulum (0.209 kg), g is the acceleration due to gravity (9.8 $m/s^2$), F is the force applied on the cart ($\pm 10$ Newtons), and $l$ is the length from the centre of mass to the pivot (0.326 m). For purposes of simulation we assume a sampling interval $\tau = 0.02$ seconds.

### B. Formulation of RL model for the Cart-pole Problem

The objective of the RL agent is to control the plant (1)-(2) using an appropriate sequence of control signals (actions) in order to balance the pendulum so that (i) the angular position of the pendulum remains within $\pm 12°$ from the upright position, and (ii) the linear position of the cart is within $\pm 2.4$ meters from the centre of the track.

The state variables $[\theta, \dot{\theta}, x, \dot{x}]$ are quantized and separated into multiple bins. A box is defined as an n-tuple comprising of one bin from each of the four state variables. Two functions *getBox* and *getBox2* have been defined with different quantization levels: The first referred to as *getBox* comprises of 162 boxes when quantized into 15 bins in the following manner: $\theta: [-12, -6), [-6, -1), [-1,0), [0,1), [1,6), [6,12]$; $\dot{\theta}: (-\infty, -50), [-50, 50], [50, \infty)$; $x: [-2.4, -0.8), [-0.8, 0.8], (0.8, 2.4]$, and $\dot{x}: (-\infty, -0.5), [-0.5, 0.5], (0.5, \infty)$. The second referred to as *getBox2* comprises of 324 boxes when quantized into 21 bins: $\theta: [-12, -6], (-6, -4], (-4, -3], (-3, -2], (-2, -1], (-1, 0], (0, 1], (1, 2], (2, 3], (3, 4], (4, 6]$ , $(6, 12]$; $\dot{\theta}: (-\infty, -50), [-50, 50], [50, \infty)$; $x: [-2.4, -0.8), [-0.8, 0.8], (0.8, 2.4]$ $\dot{x}: (-\infty, -0.5), [-0.5, 0.5], (0.5, \infty)$. If at any time the state does not belong to these discrete bins, then a failure is said to have occurred and a reinforcement signal of value -1 is generated, which marks the end of an episode. Each time a failure occurs, the pendulum is reset to the upright unstable equilibrium position, which is the initial position for the next episode. (As seen later, with the integration of swing-up with RL, such manual resetting is not required.)

Markov decision processes (MDP) play an important role in RL. They provide a mathematical framework and enables decision making in situations where outcomes are partly random and partly under the control of the agent. The core problem of MDPs is to find a policy for the agent based on its current state. The cart-pole problem can also be described using an MDP: It consists of (i) S, a finite set of states: $\theta, \dot{\theta}, x, \dot{x}$; (ii) A, a finite set of actions: Moving the cart LEFT or RIGHT by applying a force [-F, F]; (iii) P, a state transition probability matrix, $P(a, s, s') = P[S(t+1) = s'|S(t) = s, A(t) = a]$. Since the environment is deterministic, taking an action $a$ from state $s$ would always result in the next state s'. Thus, the state transition probability matrix (STPM) simplifies to the state transition matrix (STM). STM can be defined based on the dynamics of the system, given by (1) and (2); (iv) R, a reward function $R(a, s) = E[R(t+1)|S(t) = s, A(t) = a]$. A reward of -1 is given to the system whenever the state cross the restrictions defined earlier, and a reward of zero is awarded otherwise; and (v) $\gamma$, a discount factor in the range [0,1] which defines the dependency of future rewards on the current action and current state. The discount factor is varied according to the algorithm used.

In the cart-pole problem, like many other complex control problems, complete knowledge of the MDP cannot be obtained, as the state transition probabilities cannot be determined before achieving optimal control. This is because, the RL agent learns by taking random actions, with no knowledge of correct actions, until the optimal policy has been obtained. Hence, model-free RL algorithms such as Temporal Difference, Policy-gradient actor-critic and Value function approximation have been used to solve the problem. These are discussed in the next Section.

## III. MODEL-FREE LEARNING

The class of algorithms that are used to solve MDPs without the knowledge of the reward function, $R_s^a$ or the STPM, $P_{ss'}^a$ are termed as model-free methods [9]. They rely on MDPs to sample various sizes of experiences from the environment depending on the algorithm used. These experiences are then used by an agent to directly make decisions and take actions in the environment. Algorithms for model-free learning can be categorized into various types based on the length of backups, which is the number of steps for which the RL Agent interacts with the environment before updating its estimate for the quality of a state. These types are: one-step backup or temporal difference or TD(0) learning, infinite-step or full-length backup or Monte Carlo learning, N-step backup, and multi-step backup or TD(λ). Model-free methods largely use the generalized policy iteration (GPI), initially developed for dynamic programming [10] to evaluate the environment and select actions.

When using model-free learning for control, policy evaluation and policy improvement can follow either the same or different approaches in GPI. This deals with the exploration-exploitation trade-off that exists in most MDPs. This trade-off is handled by two methods: (1) On-policy control where the agent uses the same policy algorithm to select actions from the action value estimates as well as generate action value estimates, and (2) Off-policy control where the agent uses one policy algorithm, usually a greedy policy, to select actions from action value estimates, and another policy algorithm, usually an exploratory policy, to generate action value estimates.

## A. Off-Policy TD Control Algorithm (Q-Learning)

Temporal difference (TD) learning is a class of RL algorithms which involves one-step updates of the value function and bootstrapping to estimate the quality of a state. In TD methods, the quality of the state at every step is updated using reward obtained at that step and an old estimate of quality of the next state [9]. Q-Learning is a TD algorithm applied to control problems using off-policy method since two different policies are utilized by the agent. The policy used to select actions using the state-action values is the greedy policy, given by $\max_{a'} Q(S_{t+1}, a')$. On the other hand, usually an exploratory policy is used to generate the action-value estimates. The Q-learning update equation is given by:

$$Q(S_t, A_t) = Q(S_t, A_t) + \alpha[R_t + \gamma \max_{a'} Q(S_{t+1}, a') - Q(S_t, A_t)] \quad (4)$$

where $(S_t, A_t)$ corresponds to the current state and action, $(S_{t+1}, a')$ corresponds to the next state and action, $Q(S_t, A_t)$ is the quality of the agent being in state $S_t$ and taking action $A_t$, and $R_t$ is the reward obtained by taking an action $A_t$ from state $S_t$. Each step of Q-Learning updates a state-action pair: (1) The current state, $S_t$ or the internal state of the cart-pole dynamics represented in a form that the RL agent can interpret. (2) The action $A_t$ to be taken at state $S_t$. In the case of cart-pole balancing, action can be either a constant acceleration of the cart towards LEFT or RIGHT in the track.

The RL Agent is punished with a reinforcement signal $R_t$ of -1 if either the pole or the cart exceeds its limitations defined in the objective and $R_t$ of 0 otherwise. Using the update equation, Q-learning can be used to solve the cart-pole balancing problem with the following algorithm:

- Initialize all $Q(S, A) \ \forall \ S \in \mathbf{S}, \ A \in \mathbf{A}$
- For each episode:
  - For each step in an episode:
    - Given current state $S_t$, choose $A_t$ using $A_t = \mathrm{argmax}_A Q(S_t, A)$
    - Take the action $A_t$
    - Observe $R_t$ and $S_{t+1}$ from the environment
    - Update the Action value function, $Q(S_t, A_t)$ to the Q-target $R_t + \gamma \max_{a'} Q(S_{t+1}, a')$ using the update equation
    - Until the terminal state, where the state $S_t$ exceeds the limits set by the objective.

Since the state space of the cart-pole MDP is explored even with a greedy policy, the two policy algorithms followed by off-policy control are chosen as greedy policies. As a result of this selection, this special case of off-policy method converges to an on-policy approach.

## B. Actor-critic policy gradient method

### 1) Policy Gradient

The RL algorithms considered so far involve a value function which quantifies the significance of the system being the current state and taking the specified action. Action selection using these value function based methods can be performed either deterministically or with some stochasticity that can be reduced to a deterministic form based on the overall performance of the agent [9].

In Policy gradient methods, the process of action selection at every step is stochastic. It is based on the probability of selection of a particular action in each state, given by:
$$\pi_\theta(s, a) = \mathbf{P}[a \mid s, \boldsymbol{\theta}] \quad (5)$$
This can be useful in many applications where determining the accurate value function is complex. In case of cart-pole balancing problem, one such example is the upright state, where the pole is in the upright position but the agent must take either action defined by the objective. In this context, the agent may not prefer a deterministic action as it may limit the exploration across the state space.

The cost function to find the policy $\pi_\theta(s, a)$ is:
$$J_R(\boldsymbol{\theta}) = \sum_s d^{\pi_\theta}(s) \sum_a \pi_\theta(s, a) R_s^a \quad (6)$$
where $d^{\pi_\theta}(s)$ is the distribution of the MDP for $\pi_\theta$. Using stochastic gradient descent to minimize this cost function with respect to $\boldsymbol{\theta}$, and manipulating the equation:
$$\nabla_\theta \pi_\theta(s, a) = \pi_\theta(s, a) \cdot \frac{\nabla_\theta \pi_\theta(s,a)}{\pi_\theta(s,a)}$$
$$= \pi_\theta(s, a) \cdot \nabla_\theta \log(\pi_\theta(s, a)) \quad (7)$$

The expression $\nabla_\theta \log(\pi_\theta(s, a))$ is known as the score function. To represent action selection as a probability over the state, Softmax policy is chosen as the score function:
$$\nabla_\theta \log(\pi_\theta(s, a)) = x(s, a) - \mathbf{E}[x(s, \cdot)] \quad (8)$$
where $x(s, a)$ is the feature vector.

### 2) Actor-critic method

The high variance observed in policy gradient methods is a drawback. To overcome this drawback, Actor-critic method was proposed in [6], [11]. Two networks are proposed in [6] - action network and critic network. The action network learns to select actions as a function of the cart-pole system states. It consists of a single neuron having two possible outputs, +Force or –Force. The probability of generating each action depends on the box in which the system is in. Initial values of weights are zero, making the two actions equally probable. Weights are incrementally updated, and thus, the action probabilities, after receiving non-zero reinforcements which are obtained as a feedback upon failure.

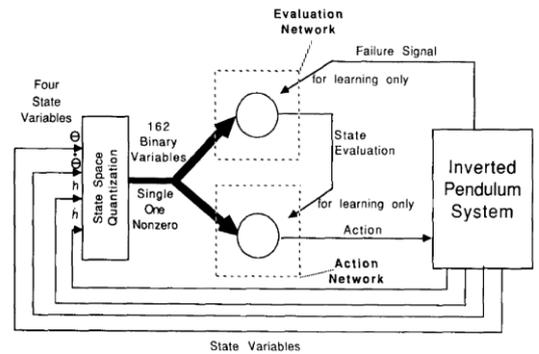

Fig. 2. Actor Critic Neuron Like elements

The critic network provides an association between the state vectors and the failure signal. The evaluation network also consists of a single neuron. This network learns the expected value of a discounted sum of future failure signals by means of TD learning [6]. Through learning, the output of the evaluation network will predict how soon a failure can be expected to occur from the current state. This prediction acts like a feedback for the action network, which enables it to learn to select a correct action when it is in a particular state.

By estimating the action value function using a critic, the high variance of the policy gradient method can be reduced. Thus, policy-gradient actor-critic methods are considered with the following parts [12]: The critic which updates action-value function $Q(S_t, A_t)$ or their parameters, and the actor which updates the policy parameters $\boldsymbol{\theta}$ in the direction of the action-value function as estimated by the critic.

The update equations for the actor network parameters are given by:

$$\theta_i(t+1) = \theta_i(t) + \alpha\big(r(t+1) + \gamma p^s(t+1) - p^s(t)\big)\bar{e}_i(t) \quad (9)$$

where $e_i(t) = \left(y(t) - \frac{1}{2}\right)x_i(t)$, $\bar{e}_i(t) = (1-\delta)\bar{e}_i(t-1) + \delta e_i(t)$, with $\bar{e}_i(0) = 0$ for $0 < \delta < 1$, and $y(t) = \begin{cases} 1, & \text{if } (s(t) + \eta(t) > 0) \\ 0, & \text{otherwise} \end{cases}$, $\eta(t)$ is a random variable chosen from a normal distribution with zero mean and standard deviation 0.1, $s(t) = \sum_{i=1}^{n} \theta_i(t) \cdot x_i(t)$ and $p^s(t) = \sum_{i=1}^{n} \theta_i(s) \cdot x_i(t)$, $p(t) = r(1)$.

The update equation for the critic network parameters are given by:

$$w(t+1) = w_i(t) + \beta\big(r(t+1) + \gamma p^s(t+1) - p^s(t)\big)\bar{x}_i(t) \quad (10)$$

where, $\bar{x}_i(t) = (1-\delta)\bar{x}_i(t-1) + \delta x_i(t)$, with $x_i(0) = 0$ for $0 < \delta < 1$, $p^s(t) = \sum_{i=1}^{n} \theta_i(s) \cdot x_i(t)$, $p(t) = r(1)$. By varying the parameters, $\alpha, \beta, \delta, \gamma, \eta$, implementing the update equation in the GPI, and testing on the cart-pole balancing problem, different results are obtained and compared in this paper.

## C. TD with Value Function Approximation

The results obtained by using the above RL methods on the cart-pole problem have a major drawback. The assumption of a discretized state space increases the effect of the value of constant force action on the performance of the system. Also, the constant force must be chosen to ensure that the system changes its state at the end of every time step. In other words, the constant force should cause $P_{ss}^a = 0$. A continuous state space is considered in order to overcome these drawbacks, and thus, a continuous-state MDP is considered [9]. Here, the four state variables are treated as continuous-time variables, $x(t) \in \mathbf{R}^4$.

### 1) Value function approximation

With continuous-state MDP, it is not possible to update the value of every state individually, as each state $x(t) \in \mathbf{R}^4$. Also, storing a separate value to represent the quality of each state result in a very large MDP, which cannot be stored in the memory efficiently. An approximate value function is considered to generalize the values from the states visited by the agent to unvisited states in the neighbourhood. This function uses a set of parameters $\boldsymbol{w} \in \mathbf{R}^4$, which can represent the quality of all states without taking up large memory space, and is defined as $\hat{q}(s, a, \boldsymbol{w}) \to q(s, a)$ where $\hat{q}(s, a, \boldsymbol{w})$ is the approximate form of the state-action value function $q(s, a)$ used in the control problem, with parameter $\boldsymbol{w}$.

### 2) Stochastic gradient descent

The goal of the RL agent defined in terms of value function approximation is to update the approximate value function towards the Q-target in case of the control problem. The parameters are updated by using a cost function $J(\boldsymbol{w})$ which represents the error between the Q-target and the approximate value and is minimized using Stochastic Gradient Descent:

$$J(\boldsymbol{w}) = \mathbf{E}\left[\big(R_{t+1} + \gamma \hat{q}(S_{t+1}, A_{t+1}, \boldsymbol{w}) - \hat{q}(S_t, A_t, \boldsymbol{w})\big)^2\right] \quad (11)$$

$$\Delta \boldsymbol{w} = -\frac{1}{2}\alpha \nabla_w J(\boldsymbol{w}) = \alpha\big(R_{t+1} + \gamma \hat{q}(S_{t+1}, A_{t+1}, \boldsymbol{w}) - \hat{q}(S_t, A_t, \boldsymbol{w})\big)\nabla_w \hat{q}(S_t, A_t, \boldsymbol{w}) \quad (12)$$

Linear combination of features can be easily applied to this problem to derive at a near-perfect RL controller:

$$\hat{q}(S, A, \boldsymbol{w}) = x(S, A)^T \boldsymbol{w} = \sum_{j=1}^{4}\big(x_j(S, A) \cdot w_j\big) \quad (13)$$

The gradient of (13) is $\nabla_w \hat{q}(S_t, A_t, \boldsymbol{w}) = x(S_t, A_t)$. Thus, the parameter update, in order to minimize the error between the Q-target and the linear value function approximate, is reduced to:

$$\Delta \boldsymbol{w} = -\frac{1}{2}\alpha \nabla_w J(\boldsymbol{w}) = \alpha(R_{t+1} + \gamma (x(S_{t+1}, A_{t+1})^T \boldsymbol{w}) - x(S_t, A_t)^T \boldsymbol{w})x(S_t, A_t) \quad (15)$$

The algorithm applied to the cart-pole problem is:
- Initialize all $\hat{q}(S, A, \boldsymbol{w}) \; \forall \; S \in \mathbf{S}, \; A \in \mathbf{A}, \boldsymbol{w}$
- For each episode:
  - For each step in an episode:
    - Given current state $S_t$, choose $A_t$ using $A_t = \text{argmax}_A \hat{q}(S, A, \boldsymbol{w})$
    - Take the action $A_t$
    - Observe $R_{t+1}$ and $S_{t+1}$ from the environment
    - Update the action value function, $\hat{q}(S, A, \boldsymbol{w})$ to the TD target using the update equation $\Delta \boldsymbol{w} = \alpha\big(R_{t+1} + \gamma \hat{q}(S_{t+1}, A_{t+1}, \boldsymbol{w}) - \hat{q}(S_t, A_t, \boldsymbol{w})\big)x(S_t, A_t)$
    - Until the terminal state, where the state $S_t$ exceeds the limitations defined by the objective.

## IV. SWING-UP AND STABILIZATION BY CLASSICAL CONTROL

This section considers a solution to perform the entire control of an inverted pendulum, from rest position at the stable equilibrium of the pendulum to a balanced upright position at the unstable equilibrium. The energy method is utilized to achieve this transfer. The control is switched to one based on LQR in order to stabilize the pendulum near the upright unstable equilibrium position. The swing-up and stabilization strategies are integrated to perform the full control of the inverted pendulum. This is done by switching the controllers from swing-up to stabilization when the pendulum angle is within $\pm 12°$ and back to swing-up when the angle exceeds $\pm 12°$, as suggested in the cart-pole MDP.

## A. Swing-up using Energy Control Method

Equation (1) is considered to derive the swing-up strategy with the following modification [13]: $\bar{I}_p = I_p + ml^2$; $I_p = \left(\frac{4}{3}\right)(M+m)l^2$. The total energy of the pendulum at any state is:

$$E = \frac{1}{2}\bar{I}_p\dot{\theta}^2 + mgl(\cos\theta - 1) \qquad (16)$$

With change in energy depending on $\dot{\theta}\cos\theta$ and using the Lyapunov function defined by $= (E - E_0)^2/2$, the control law required to reach the target energy $E_0$ is given by:

$$u = k(E - E_0)\dot{\theta}\cos\theta \qquad (17)$$

The above method does not consider the restrictions of the finite cart track length. Introducing the restrictions [14], the Lyapunov function is modified to $V = \frac{1}{2}((E - E_0)^2 + ml\lambda\dot{x}^2)$ and thus, the new control law is:

$$u = k((E - E_0)\dot{\theta}\cos\theta - \lambda\dot{x}) \qquad (18)$$

where $\lambda$ is a parameter to restrict the linear motion of the cart.

## B. Stabilization using Linear Quadratic Regulator

When the pendulum reaches a position of $\pm 12°$ from the upright position using swing-up the control is switched from swing-up controller to the stabilization controller. The state space near the upright position of the pendulum ensures small values of $\theta$. Thus, the model can be linearized and a robust linear controller such as the LQR can be utilized to stabilize the pendulum near this position. The linearized state space model is given by $\dot{X} = AX + Bu$, where $X = [\theta, \dot{\theta}, x, \dot{x}]^T$

$$A = \begin{bmatrix} 0 & 1 & 0 & 0 \\ 0 & \dfrac{I_p}{(M+m+m^2l^2)} & \dfrac{m^2l^2g/I_p}{(M+m+\frac{m^2l^2}{I_p})} & 0 \\ 0 & 0 & 0 & 1 \\ 0 & \dfrac{b}{\left(\dfrac{(M+m)I_p}{ml}\right)+ml} & \dfrac{(M+m)g}{\left(\dfrac{(M+m)I_p}{ml}\right)+ml} & 0 \end{bmatrix}$$

$$B = \begin{bmatrix} 0 \\ \dfrac{I_p}{(M+m+m^2l^2)} \\ 0 \\ \dfrac{-1}{\left(\dfrac{(M+m)I_p}{ml}\right)+ml} \end{bmatrix}$$

The state-feedback stabilizing control is $u = -KX$. Thus, $\dot{X} = (A - BK)X$, where the LQR gain $K = R^{-1}B^TP$ is obtained from the solution of the corresponding Algebraic Riccati Equation (ARE) $A^TP + PA - PBR^{-1}B^TP + Q = 0$ for specified Q and R. Here, $Q = C^TC$ and $R = 1$, where, $C = \begin{bmatrix} 1 & 0 & 0 & 0 \\ 0 & 0 & 1 & 0 \end{bmatrix}$. For the example system considered here, $K = [-1.0000 \quad -1.7788 \quad -26.3106 \quad -3.8440]$.

## V. INTEGRATION OF SWING-UP WITH STABILIZATION USING RL

To compare the performance of the RL algorithms with the LQR strategy, a novel approach has been proposed to integrate the swing-up strategy with RL algorithms. The reasons for this are as follows: (i) The existing research on application of RL to the cart-pole problem specifies that the pendulum must be reset to the unstable equilibrium position after each failure. However, this approach when applied to a physical system seems redundant and cumbersome. Instead, the pendulum is reset to the stable equilibrium position each time failure occurs, and then swung up to a pre-defined value of angular position from where the control is switched to the RL controller. (ii) Once the failure occurs, the pendulum reaches the stable equilibrium due to natural damping, from where swing-up is initialized. So, the whole process can be automated. (iii) When the pendulum is swung up to a window of pre-defined range of angular position values, the pendulum can end up in a different state each time. Hence, the scope for exploration increases. In this paper, the stabilization controller from the previous section is replaced with the RL agent, and the switching occurs at $\pm 12°$ angle.

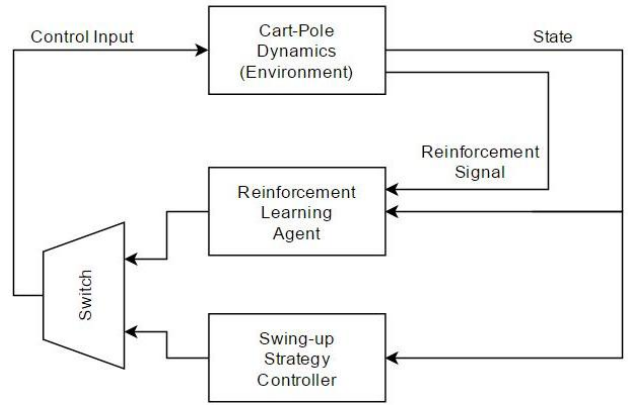

Fig. 3. Swing-up using energy method and stabilization using reinforcement learning.

## VI. RESULTS

The controller for the cart-pole system is assumed to have reached its objective when $\theta$ and $x$ remains within the bounds specified earlier for more than 100,000 steps, where each step is 0.02 seconds.

### A. Off-Policy TD Control Algorithm (Q-Learning)

Q-Learning is a basic TD control algorithm. The results with a force of {+10, -10} and parameters α=0.5 and γ=0.99, are shown in Fig. 4. Evidently, Q-Learning achieves optimal policy. However, it does so after 420 trials. The cart almost hits one end of the track during the experiment and the pendulum angle also reaches the edge of its restriction limits. Specifically, the range of the angles covered by the pendulum measured from the upright position during the experiment is [11°,+10°], and the range of the cart position is [-1.5m, +2.4m]. The performance can be improved with different choice of parameters; the results are summarised in Table I.

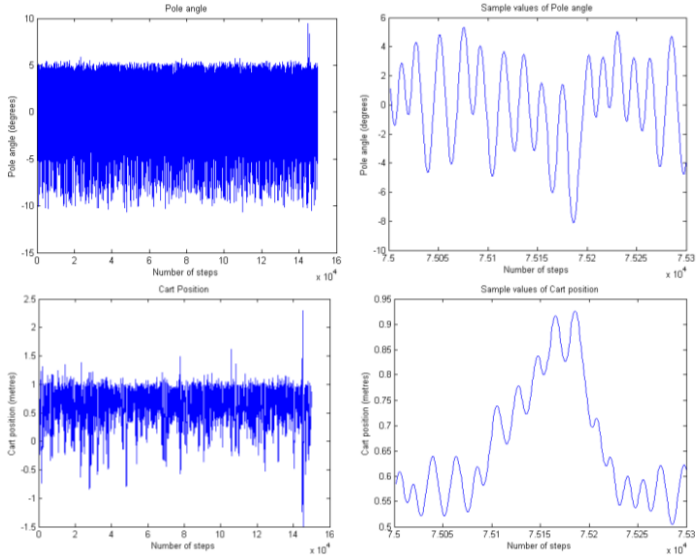

Fig. 4. Q-Learning on cart-pole problem. Force = {+10, -10}, learning rate = 0.5, discount factor = 0.99.

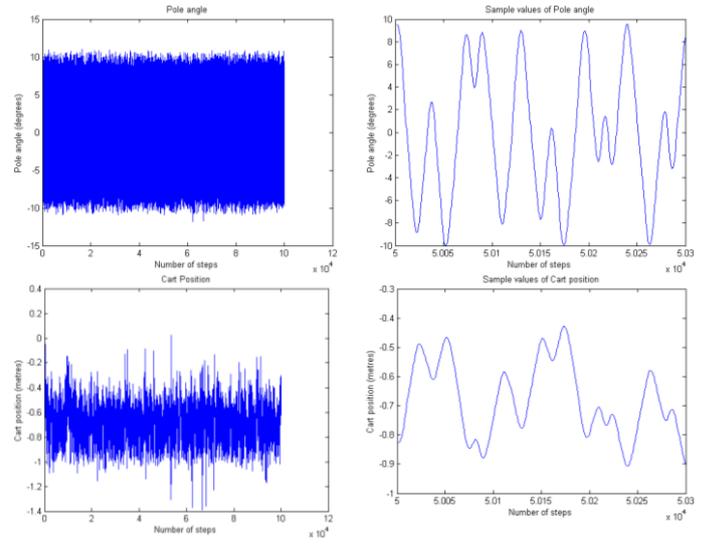

Fig. 5. Actor-critic method on cart-pole problem: Force = {+10, -10}, learning rate = 1000, discount factor = 0.95, Eligibiility trace for: actor = 0.9, critic = 0.8.

## B. Actor-critic policy gradient method

The policy gradient actor-critic method involves the calculation of a value function by the critic network as well as the selection of the best action for the given state, and the state value function, through a probabilistic approach. The chosen parameters are as follows: The force is {+10, -10}N, $\alpha$=1000, $\gamma$=0.95, $\lambda_w$=0.9, and $\lambda_v$=0.8. These parameters have been adjusted to achieve optimal policy. However, as seen from Fig. 5, this method performs rather poorly although the specifications on the cart and the pole are met.

With a large learning rate, $\alpha = 1000$, it would be expected that the agent would never learn the optimal policy. However, as can be observed from the results the optimal policy is achieved albeit with large oscillations in the steady-state implying a large expenditure of control energy. Further, the range of deviations in $\theta$ is rather high and covers nearly all of the allowable state space. Even the cart position varies widely from -1.7m to 0.2m. This again implies a rather large control effort. Furthermore, the variations in these state variables are nearly periodic which is perhaps characteristic of actor-critic methods.

## C. Temporal Difference with Value Function Approximation

The results with a linear value function approximation are shown in Fig. 6 with the parameters FORCE = {+10, -10}N, $\alpha$=0.07, $\gamma$=0.992. Evidently, this approach provides rather satisfactory results. Observe that the RL agent achieves the optimal policy in merely 19 episodes, a remarkable improvement over the earlier approaches. Further, the amplitude of oscillations is quite minimal. The angle $\theta$ lies between -0.4° and 0.3°, and the cart position in the range [-0.062, +0.054]m. The oscillations in the beginning of the episode are also minimal. Observe that when balanced, the cart has deviated from the centre.

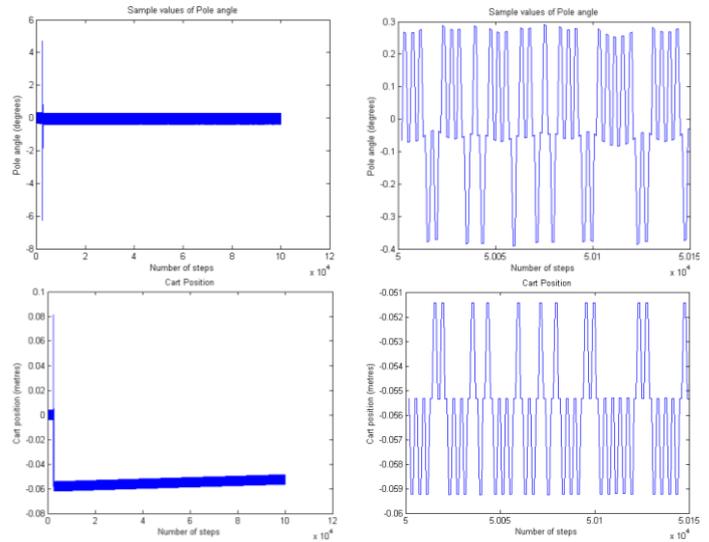

Fig. 6. Temporal Difference with Value Function Approximation on cart-pole problem Force = {+10, -10}, learning rate = 0.07, discount factor = 0.992.

## D. Classical Swing-up and LQR Stabilization

The results of this experiment are shown in Fig. 7. The swing-up controller ensures that the pole is moved from its stable position ($\theta = \pi$) to the desired position. This is carried out in a manner such that the cart never hits the boundaries of the rails, and it requires several oscillations about the stable position. The parameter $\lambda$ in the design determines the cart length limitations. Once $\theta$ reaches the specified cone, the LQR controller ensures the pole is stabilized at the desired position.

## E. Classical Swing-up and RL Stabilization

The results of the integration of the classical swing-up with the RL algorithm are shown in Fig. 8. Evidently, stabilization is independent of the swing-up, and the scope for exploration of the RL controller has been increased. Further, the system is now fully automated in that whenever the RL policy fails, the swing-up controller kicks in automatically.

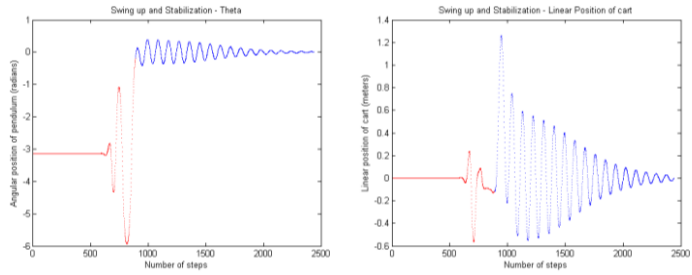

Fig. 7. Swing-up and stabilization of inverted pendulum using energy method and linear quadratic regulator

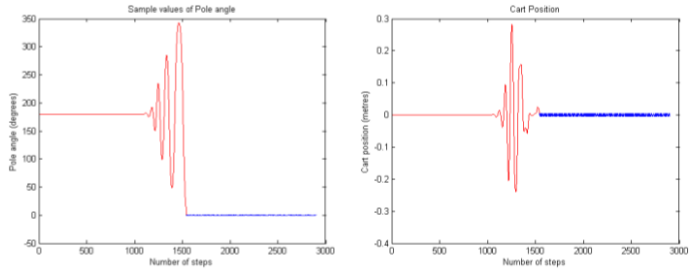

Fig. 8. Swing-up and stabilization of inverted pendulum using energy method and Value Function Approximation Force = {+10, -10}, learning rate = 0.07, discount factor = 0.992

TABLE I.  COMPARISON OF ALGORITHMS

| Algorithm | Force (in N) | $\alpha$ | $\gamma$ | State-space Quantizer | $\lambda_w$ and $\lambda_v$ | $\theta$ Range (in °) | $x$ Range (in m) | Episodes until optimal policy |
|---|---|---|---|---|---|---|---|---|
| SARSA(0) | {+10, -10} | 0.5 | 0.99 | getBox | - | [-11,+10] | [-1.5, +2.4] | 420 |
| SARSA(0) | {+15, -15} | 0.6 | 0.99 | getBox2 | - | [-2.9, +2.9] | [-1. +1] | 380 |
| SARSA(0) | {+30, -30} | 0.4 | 0.99 | getBox | - | [-3, +6.3] | [-0.1, +0.6] | 24 |
| Q-Learning | {+10, -10} | 0.5 | 0.99 | getBox | - | [-11,+10] | [-1.5, +2.4] | 420 |
| Q-Learning | {+15, -15} | 0.6 | 0.99 | getBox2 | - | [-2.9, +2.9] | [-1, +1] | 380 |
| Q-Learning | {+30, -30} | 0.4 | 0.99 | getBox | - | [-3, +6.3] | [-0.1, +0.6] | 24 |
| Value Function Approximation | {+10, -10} | 0.07 | 0.99 | - | - | [-0.4, +0.3] | [-0.062, +0.054] | 19 |
| Policy Gradient Actor-Critic | {+10, -10} | 1000 | 0.95 | 0.9 | 0.8 | [-12, +12] | [-1.7, +0.2] | 62 |
| SARSA(0)/Q with Swing Up | {+10, -10} | 0.5 | 0.99 | getBox | - | [-8.6, +11.5] | [-0.6,+0.5] | 430 |
| Policy gradient Actor-Critic with Swing Up | {+10,-10} | 1000 | 0.95 | getBox | 0.8, 0.9 | [-11,11] | [-1,1.7] | 255 |
| Value function approx with Swing Up | {+10,-10} | 0.07 | 0.99 | - | - | [-0.4, +0.3] | [-0.062, +0.054] | 19 |

CONCLUSIONS

The performances of several reinforcement learning algorithms are compared when applied to the cart-pole problem. In discrete state space, actor-critic policy gradient method has converged faster and performed better stabilization than Q-Learning. The range of angular positions of the pendulum and linear positions of the cart for the optimal policy decreases with the transition from discrete to continuous state space. Value function approximation, has shown the best performance among the three algorithms. Further, the integration of swing-up using the energy method has not affected the performance of the individual algorithms. In contrast, this integration has achieved automation.